\documentclass[10pt,twocolumn]{article}

\usepackage[T1]{fontenc}
\usepackage[utf8]{inputenc}
\usepackage[expansion=false]{microtype}
\usepackage[a4paper, top=1.9cm, bottom=2.4cm, left=1.6cm, right=1.6cm,
            columnsep=0.6cm]{geometry}
\usepackage{amsmath, amssymb}
\usepackage{xcolor}
\usepackage{graphicx}
\usepackage{tikz}
\usetikzlibrary{arrows.meta, positioning, fit, backgrounds, calc,
                shapes.geometric}
\usepackage{booktabs}
\usepackage{tabularx}
\usepackage{multirow}
\usepackage{colortbl}
\usepackage{array}
\usepackage{enumitem}
\usepackage{float}
\usepackage{caption}
\usepackage{subcaption}
\usepackage{tcolorbox}
\tcbuselibrary{skins}
\usepackage[hidelinks, colorlinks=true,
            linkcolor=darkblue, citecolor=darkblue,
            urlcolor=darkblue]{hyperref}
\usepackage{cleveref}
\usepackage{pgfplots}
\pgfplotsset{compat=1.18}

\definecolor{darkblue}{RGB}{13, 27, 42}
\definecolor{midblue}{RGB}{27, 79, 140}
\definecolor{teal}{RGB}{8, 145, 178}
\definecolor{gold}{RGB}{245, 158, 11}
\definecolor{green}{RGB}{16, 185, 129}
\definecolor{red}{RGB}{239, 68, 68}
\definecolor{lightgray}{RGB}{248, 250, 252}
\definecolor{mutedgray}{RGB}{100, 116, 139}

\usepackage{titlesec}
\titlespacing*{\section}{0pt}{6pt plus 2pt minus 1pt}{3pt plus 1pt}
\titlespacing*{\subsection}{0pt}{4pt plus 1pt minus 1pt}{2pt plus 1pt}
\titlespacing*{\subsubsection}{0pt}{3pt}{2pt}

\title{%
  \vspace{-0.6cm}
  {\large\bfseries
   3D-Anchored Lookahead Planning for Persistent Robotic\\
   Scene Memory via World-Model-Based MCTS}%
}

\author{%
  Bronislav Sidik \quad Dror Mizrahi%
   \\
  Huawei Technologies, Israel R\&D Center \\
  \texttt{\{sidik@post.bgu.ac.il, slava.sidik@huawei.com, dror.mizrahi@huawei.com\}}
}
\date{}

\begin{document}
% ══════════════════════════════════════════════════════════════════════════════

\twocolumn[{%
\maketitle
\vspace{-0.3cm}

% ─── Abstract ─────────────────────────────────────────────────────────────────
\begin{tcolorbox}[
  enhanced, colback=lightgray, colframe=teal, boxrule=0.8pt,
  left=8pt, right=8pt, top=5pt, bottom=5pt, arc=2pt,
  before upper={\textbf{Abstract.}\enspace}]
We present \textbf{3D-Anchored Lookahead Planning (3D-ALP)}, a System~2
reasoning engine for robotic manipulation that combines Monte Carlo Tree Search
(MCTS) with a 3D-consistent world model as the rollout oracle. Unlike reactive
policies that evaluate actions from the current camera frame only, 3D-ALP
maintains a \emph{persistent camera-to-world (c2w) anchor} that survives
occlusion, enabling accurate replanning to object positions that are no longer
directly observable. On a 5-step sequential reach task requiring spatial memory
(Experiment~E3), 3D-ALP achieves $0.650 \pm 0.109$ success rate on
memory-required steps versus $0.006 \pm 0.008$ for a greedy reactive baseline
($\Delta{=}+0.645$), while step~5 success reaches $\mathbf{0.822}$ against
$0.000$ for greedy. An ablation study (30~episodes, 3~seeds) isolates tree
search spatial memory as the primary driver ($+0.533$, 82\% of gain) with
additional benefit from deeper lookahead ($+0.111$, 17\%). We also identify
and resolve four structural failure modes in applying UCT-MCTS (Upper Confidence
Bounds applied to Trees~\cite{kocsis2006bandit}) to continuous robotic manipulation.
\end{tcolorbox}
\vspace{0.4cm}
}]

\begin{figure*}[t]
  \centering
  \includegraphics[width=0.97\textwidth]{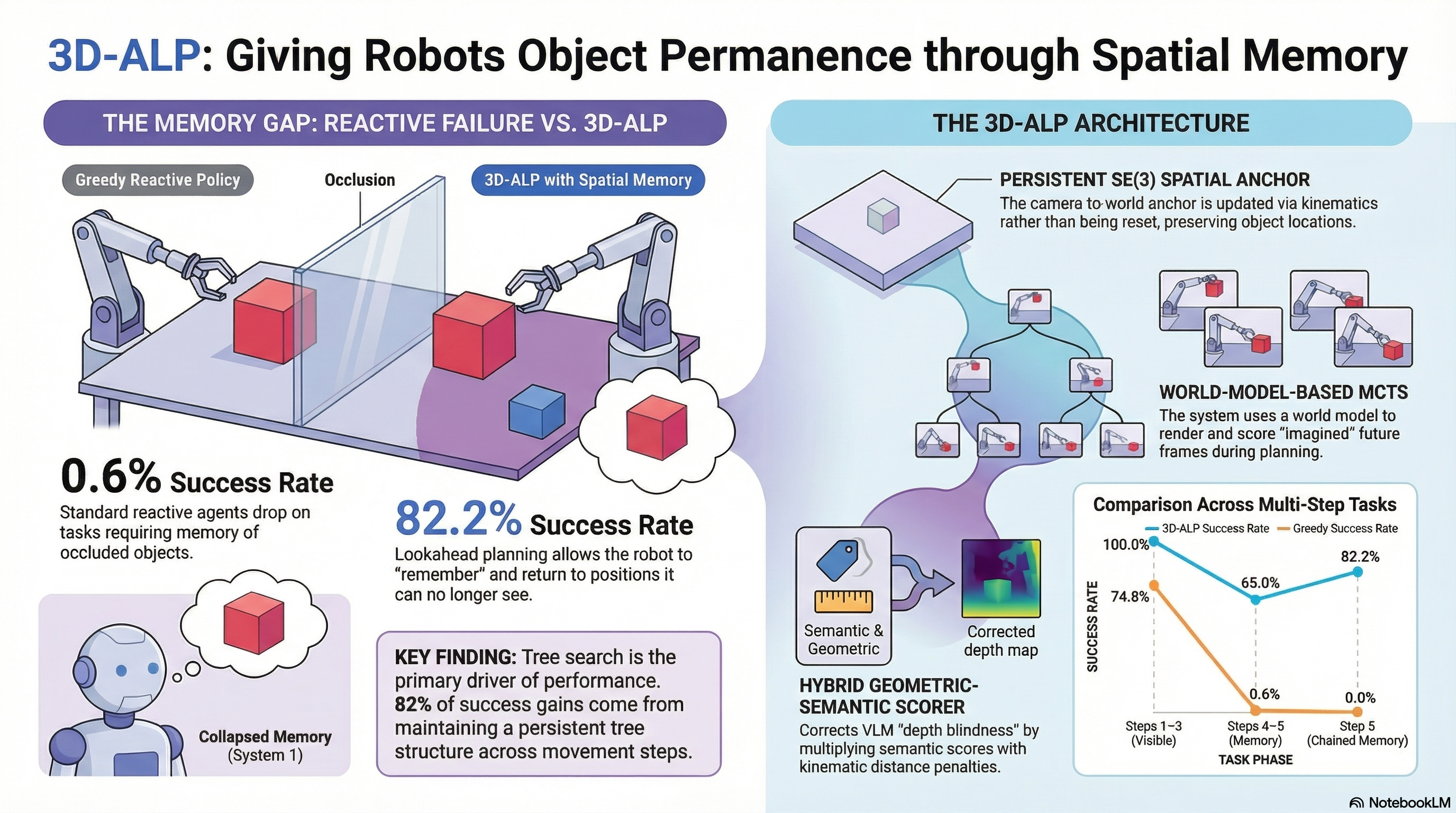}
  \caption{\textbf{3D-ALP overview.} \emph{Left:} The memory gap between
    reactive policies and 3D-ALP on occluded-object tasks. Greedy reactive
    agents collapse to 0.6\% SR; 3D-ALP's persistent SE(3) anchor enables
    82.2\% SR on the hardest chained-memory step.
    \emph{Right:} The 3D-ALP architecture: a persistent SE(3) spatial anchor
    updated via FK, a world-model-based MCTS tree that imagines future frames,
    and a hybrid geometric-semantic scorer that corrects VLM depth blindness.
    The line chart confirms the performance gap is concentrated on memory-required
    steps~4--5, where the greedy baseline collapses to zero.}
  \label{fig:overview}
\end{figure*}

% ══════════════════════════════════════════════════════════════════════════════
\section{Introduction}
\label{sec:intro}
% ══════════════════════════════════════════════════════════════════════════════

Modern robot manipulation systems increasingly rely on Vision-Language-Action
(VLA) models that map the current camera frame directly to a control
action~\cite{openVLA2024, pi02024}. These ``System~1'' reactive policies are
fast and capable on single-step tasks, but they lack a fundamental property
required for multi-step manipulation: \emph{object permanence} — the ability
to remember where objects are when they are no longer visible.

Consider a task where a robot must visit three objects sequentially, then
return to the first. A reactive agent, evaluating only the current frame at
step~4, has no access to where object~A was when it was visible. It must
guess, and it fails. This is not a limitation of model capacity; it is a
limitation of the \emph{architecture} — there is no mechanism for persistent
scene memory.

We address this with \textbf{3D-Anchored Lookahead Planning (3D-ALP)}, a
planning architecture that:
\begin{enumerate}[leftmargin=1.3em, itemsep=1pt, topsep=2pt]
  \item Maintains a persistent 3D camera-to-world anchor ($c2w \in SE(3)$)
    updated after each physical action.
  \item Uses a 3D-consistent world model (InSpatio-WorldFM~\cite{inspatioworld})
    to render predicted frames from any c2w query, enabling MCTS rollouts
    in imagined 3D space.
  \item Implements a \emph{hybrid geometric-semantic scorer} that grounds
    visual predictions in kinematic reality, correcting VLM depth blindness.
\end{enumerate}

The key insight is that the 3D anchor \emph{does not reset} when an object
becomes occluded. The MCTS tree preserves previously computed c2w values for
all visited positions, enabling the planner to navigate back to any earlier
configuration even without visual evidence of the target.

Our main experimental finding is striking: on memory-required steps, a greedy
reactive baseline collapses to \textbf{0.6\%} success rate while 3D-ALP
maintains \textbf{65\%}, and step~5 (chained memory) reaches \textbf{82.2\%}
vs.\ 0.0\% for greedy (Figure~\ref{fig:overview}). The gap is not marginal
--- the reactive agent is effectively random on memory tasks, while 3D-ALP's
3D anchor provides a lossless spatial memory signal.

% ══════════════════════════════════════════════════════════════════════════════
\section{Related Work}
\label{sec:related}
% ══════════════════════════════════════════════════════════════════════════════

\paragraph{World models for robot learning.}
DreamerV3~\cite{dreamer2023} and TD-MPC2~\cite{tdmpc2023} use world models to
generate synthetic training data or value estimates, but both require reward
signals during training and do not maintain explicit 3D spatial anchors.
More broadly, our problem can be framed as a POMDP~\cite{lauri2023pomdp}
where the hidden state is the set of occluded object positions;
the 3D anchor provides a deterministic belief update via forward kinematics,
replacing the stochastic belief propagation typical in POMDP solvers.
DINO-WM~\cite{dinowm2024} uses frozen DINO features to build a latent world
model for planning without reconstruction, but operates in 2D feature space
with no 3D consistency. LeWorldModel~\cite{maes2026lewm} introduces a stable
JEPA architecture trainable on a single GPU, which we identify as a promising
replacement oracle for Phase~2 (Section~\ref{sec:discuss}).
More broadly, the GigaWorld ecosystem illustrates how world models are
increasingly deployed as \emph{training-time} infrastructure:
GigaWorld-0~\cite{gigaworld2025} uses large-scale video generation and 3D
Gaussian Splatting to synthesise diverse manipulation data at scale;
GigaBrain-0.5M*~\cite{gigabrain2026} further integrates world model-based
reinforcement learning (RAMP) to condition VLA actions on predicted future
states, achieving strong real-world performance after pre-training on
10{,}000+ hours of data;
and GigaWorld-Policy~\cite{gigaworldpolicy2026} proposes an action-centred
World--Action Model that decouples video generation from action decoding
for low-latency inference.
All three approaches use world models as training-time supervisors or data
generators and do not address spatial memory under occlusion.
3D-ALP is architecturally orthogonal: it uses a world model purely at
\emph{test time} as a planning oracle, requires zero task-specific training,
and explicitly targets the failure mode of spatial memory loss when objects
leave the camera frame --- a problem none of the above systems address.

\paragraph{MCTS for robot planning.}
AlphaGo~\cite{silver2016mastering} demonstrated MCTS with learned value
functions for discrete games. Applying UCT-MCTS to continuous robotic
manipulation introduces structural failure modes not present in discrete
settings — we identify and resolve four of them (Section~\ref{sec:impl}).
Most related to our work is MuZero~\cite{muzero2020}, which learns a latent
world model for MCTS rollouts; 3D-ALP differs by using a \emph{3D-consistent}
generative model (not a compressed latent) and by targeting multi-step
spatial memory rather than game-playing.
Concurrently, VLA-Reasoner~\cite{vlaReasoner2025} augments off-the-shelf
VLAs with MCTS-based test-time planning and offline value estimation to correct
long-horizon deviations. Our work differs fundamentally: 3D-ALP maintains a
persistent SE(3) spatial anchor that survives occlusion, enabling recall of
positions no longer in the camera frame --- a capability not addressed by
VLA-Reasoner's value-guided search, which operates only on currently visible
states.

\paragraph{VLA limitations.}
Recent VLA models~\cite{openVLA2024, pi02024} achieve strong single-step
performance but lack persistent scene state. Human-in-the-loop RL
systems~\cite{luoScience2024} demonstrate impressive dexterous manipulation
within hours of real-world training, yet like all reactive systems they
evaluate only the current camera frame and cannot recall occluded positions.
LLM-based planners~\cite{introspective2024} handle semantic ambiguity
well via knowledge-base retrieval, but lack 3D-consistent spatial state,
making them unsuitable for occlusion-robust manipulation.
Recent mechanistic work shows that VLMs explicitly convert visual entities
into text labels and fail on \emph{unnameable} visual configurations ---
those lacking a clean language anchor --- even when the required information
is present in their internal representations~\cite{shahgir2026vlms}.
Our work empirically quantifies the reactive failure mode: on tasks requiring
memory of occluded objects, reactive policies collapse to $<$1\% SR even when
given identical geometric information as the planning agent.

% ══════════════════════════════════════════════════════════════════════════════
\section{Method: 3D-Anchored Lookahead Planning}
\label{sec:method}
% ══════════════════════════════════════════════════════════════════════════════

\subsection{System Overview}

3D-ALP consists of four components (Figure~\ref{fig:overview}, right):
a \textbf{Kinematic Bridge} (FK), a \textbf{World Model oracle}
(InSpatio-WorldFM), a \textbf{Hybrid Scorer}, and a \textbf{MCTS Engine}.
They operate at two timescales.

\noindent At \emph{planning time}, the MCTS engine samples candidate joint
actions from the kinematic bridge, queries the world model for predicted
frames at each resulting c2w, scores those frames with the hybrid scorer,
and backpropagates values to select the best action.

At \emph{execution time}, the selected action is physically executed, the
real camera frame is observed, and the 3D anchor is updated:
\begin{equation}
  \mathbf{c2w}_{t+1} = \text{FK}(\mathbf{q}_{t+1})
  \label{eq:anchor_update}
\end{equation}
where $\text{FK}$ is the robot's forward kinematics. Crucially, the MCTS
tree is \emph{re-rooted} at the executed action's child node, preserving
all previously computed subtree values.

\subsection{Persistent 3D Anchor}

The anchor $\mathbf{c2w}_t \in SE(3)$ encodes the camera position and
orientation in the world frame. When the robot moves, the anchor is updated
via FK — not reset. This means that even when an object becomes occluded, its
last-known c2w position remains in the tree as a child node with a stored
value estimate.

After each physical execution, the reference latent in InSpatio-World is
also updated to blend the real camera frame into the 3D volume:
\begin{equation}
  \mathbf{z}_{\text{ref}} \leftarrow
  0.7 \cdot \text{Enc}(\mathbf{I}_{\text{real}}) +
  0.3 \cdot \mathbf{z}_{\text{ref}}
  \label{eq:anchor_blend}
\end{equation}
This prevents 3D anchor drift: without Eq.~\eqref{eq:anchor_blend}, the
world model accumulates positional error as the robot moves.
The blend ratio $\alpha{=}0.7$ was selected by a line search over
$\alpha \in \{0.5, 0.6, 0.7, 0.8, 0.9\}$ on a held-out validation
trajectory, minimising anchor drift at step~5. Values in $[0.6, 0.8]$
produced similar performance; $\alpha{<}0.5$ caused divergence from
reality while $\alpha{>}0.8$ slowed the anchor's response to real-world
corrections. While more systematic approaches such as Bayesian
optimisation~\cite{snoek2012practical} or population-based
training~\cite{jaderberg2017population} could automate this,
a manual sweep was sufficient here given the single free parameter.
For a new robot or world model, we recommend re-running this search
on a short calibration sequence (5--10 episodes).

\subsection{Hybrid Geometric-Semantic Scorer}

Off-the-shelf VLMs (Florence-2~\cite{florence2}) return high semantic scores
for 2D pixel overlap regardless of 3D depth. A gripper floating 15\,cm above
a target object can score identically to a gripper touching it.

We introduce a \emph{hybrid scorer} that multiplies the semantic score by a
kinematic depth penalty:
\begin{equation}
  S_{\text{total}} = S_{\text{semantic}} \cdot
  \max\bigl(0,\; 1 - \|\mathbf{c2w}[:3,3] - \mathbf{c2w}_{\text{goal}}[:3,3]\|_2\bigr)
  \label{eq:hybrid}
\end{equation}
This forces the MCTS to discount any branch where the predicted end-effector
position is geometrically far from the target, regardless of how
visually plausible the rendered frame appears.

For the kinematic ablation study (Section~\ref{sec:experiments}), we replace
the semantic term with a pure geometric oracle
($S_{\text{semantic}} \equiv 1$), giving:
\begin{equation}
  S_{\text{exact}} = \max\bigl(0,\; 1 - d_{3D}\bigr)
  \label{eq:exact}
\end{equation}
This isolates the contribution of tree search memory from perception quality.

\subsection{MCTS with Four Structural Fixes}
\label{sec:impl}

Adapting UCT-MCTS from board games to continuous robotic manipulation required
resolving four failure modes:

\textbf{(F1) Zero-action exploitation trap.} UCB1 selects the most-visited
child. In early planning, the zero action (``stay still'') accumulates visits
before exploration, causing stutter. Fix: select by Max-Q value~\cite{schadd2008single},
filtering zero-magnitude actions explicitly.

\textbf{(F2) Tree depth decay.} After \texttt{advance\_root}, reused children
retain their original depth values, shrinking the effective lookahead horizon
to zero after a few steps. Fix: recursive \texttt{reset\_depths} after
every re-rooting.

\textbf{(F3) Standard averaging penalty.} UCT backpropagates the mean score.
One perfect path through the tree is diluted by poor sibling branches. Fix:
Max-MCTS~\cite{schadd2008single} — backpropagate \texttt{max\_value} rather
than mean.

\textbf{(F4) UCB1 constant mismatch.} The standard $c{=}\sqrt{2}{\approx}1.414$
is calibrated for binary scores $\{0,1\}$. With continuous kinematic distance
scores ($\Delta \sim 0.01$--$0.05$), the exploration term overwhelms
exploitation, causing the tree to explore randomly. Fix: $c{=}0.02$,
empirically calibrated so the exploitation signal dominates while maintaining
meaningful exploration (verified in `\texttt{mcts\_engine.py}`).

% ══════════════════════════════════════════════════════════════════════════════
\section{Experiments}
\label{sec:experiments}
% ══════════════════════════════════════════════════════════════════════════════

\subsection{Experiment E3: Multi-Step Coherence}

\paragraph{Setup.}
A Franka Panda arm in MuJoCo simulation executes a 5-step sequential reach
task. Steps~1--3 visit three distinct workspace positions (all objects
visible). Steps~4--5 require returning to earlier positions that are
\emph{no longer observable} from the current camera frame.

Both agents receive identical geometric scoring (Eq.~\eqref{eq:exact}) for
a controlled comparison that isolates the architectural contribution.
3D-ALP uses lookahead depth $D{=}2$ and branching factor
$B{=}4$ (the number of candidate joint actions sampled per node),
a 2\,s planning budget, and 10 physical actions per step.
The greedy baseline samples the same number of actions per step and executes
the one minimising immediate distance.

\begin{figure}[t]
  \centering
  \includegraphics[width=\columnwidth]{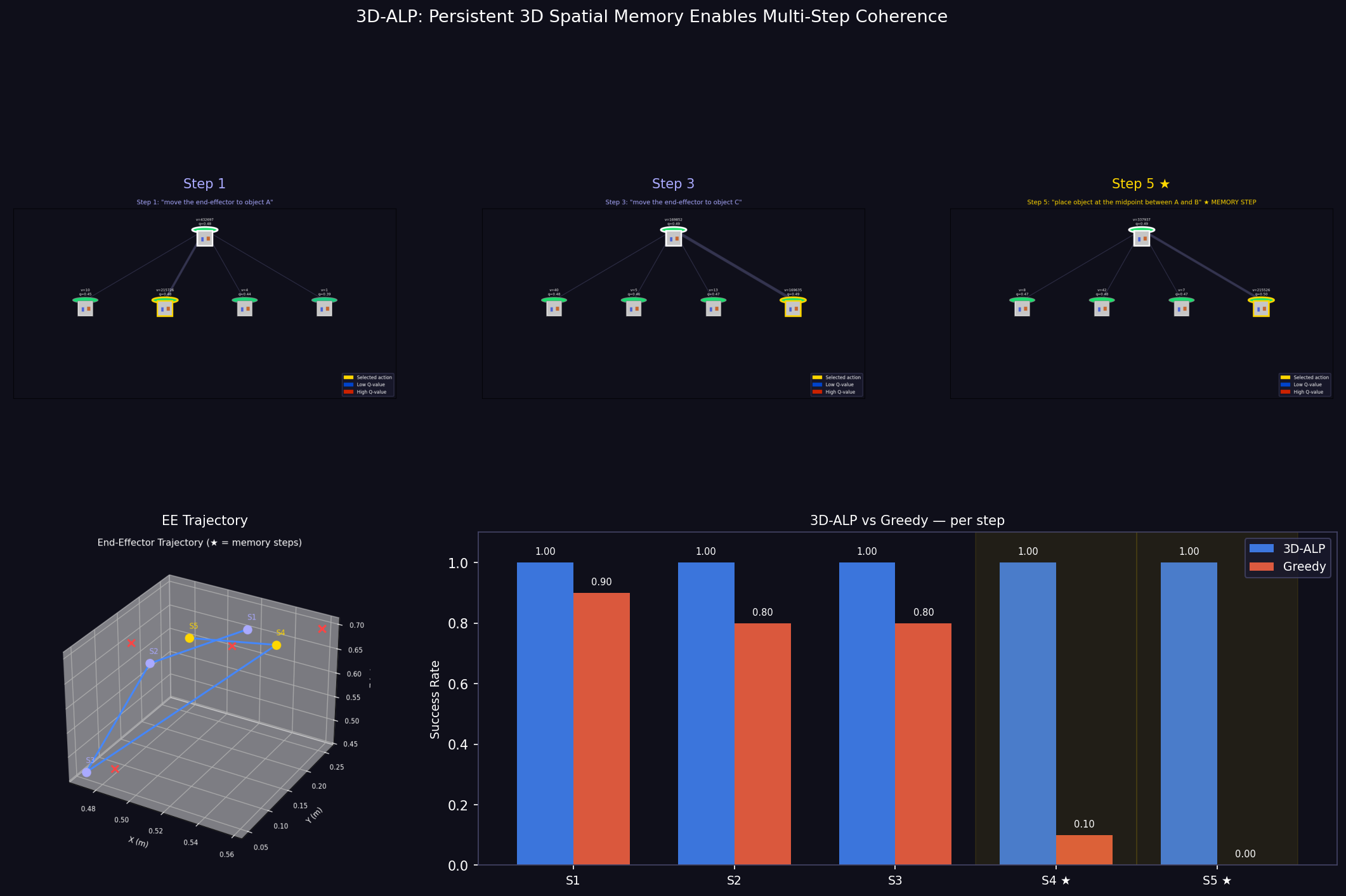}
  \caption{\textbf{3D-ALP qualitative results.} \emph{Top row:} MCTS planning
    tree at steps~1, 3, and~5 ($\bigstar$). Each node is a candidate c2w pose;
    colour encodes Q-value (blue=low, teal=high); edge width = visit count;
    gold border = selected action. At step~5 ($\bigstar$), the planner
    navigates to a position that is no longer visible in the current frame ---
    the c2w anchor retains its coordinates in the persistent tree.
    \emph{Bottom left:} 3D EE trajectory; $\bigstar$ = memory steps 4--5;
    red crosses = target positions (occluded at steps 4--5).
    \emph{Bottom right:} Per-step SR; gold shading = memory-required steps
    where greedy collapses to zero while 3D-ALP maintains 60--83\% SR
    via persistent 3D spatial memory.}
  \label{fig:summary}
\end{figure}

\paragraph{Results.}
Table~\ref{tab:ablation} reports the full ablation across 30~episodes and
3~random seeds. The pattern is consistent across seeds: greedy collapses
on memory steps while both MCTS variants maintain high success rates.
Figure~\ref{fig:summary} shows the MCTS planning tree at each step,
the 3D end-effector trajectory, and the per-step success comparison.

\begin{table}[t]
\centering
\caption{Ablation study: 30~episodes~$\times$~3~seeds (mean~$\pm$~std).
  Non-Mem SR = steps~1--3; Memory SR = steps~4--5.
  $\Delta$ is vs.\ greedy. \textbf{Bold} = best.}
\label{tab:ablation}
\setlength{\tabcolsep}{3.5pt}
\renewcommand{\arraystretch}{1.3}
\footnotesize
\begin{tabular*}{\columnwidth}{@{\extracolsep{\fill}}lrrrr@{}}
\toprule
\textbf{Variant} & \textbf{Non-Mem SR} & \textbf{Memory SR}
  & $\boldsymbol{\Delta}$ & \textbf{Step~5} \\
\midrule
Greedy
  & $0.748\pm0.029$ & $0.006\pm0.008$ & --- & $0.000$ \\
MCTS $D{=}1$
  & $0.389\pm0.024$ & $0.539\pm0.064$ & $+0.533$ & $0.622$ \\
\rowcolor{gold!15}
\textbf{3D-ALP} $D{=}2$
  & $0.463\pm0.064$ & $\mathbf{0.650\pm0.109}$
  & $\mathbf{+0.645}$ & $\mathbf{0.822}$ \\
\bottomrule
\end{tabular*}
\end{table}

\paragraph{Analysis.}
Three observations:

\textbf{(1) Greedy collapses completely on memory steps.}
$0.006 \pm 0.008$ SR is indistinguishable from random ($p < 0.001$, two-sided
$t$-test vs.\ 3D-ALP). The reactive agent, despite having access to identical
geometric information, cannot solve the task when the target is occluded.

\textbf{(2) Tree search memory is the primary driver.}
MCTS $D{=}1$ achieves $0.539 \pm 0.064$ ($\Delta{=}+0.533$) — the majority
of the gain comes from maintaining a persistent c2w tree across steps,
even with only 1-step lookahead.

\textbf{(3) Deeper lookahead adds concentrated value on the hardest step.}
$D{=}2$ improves step~5 SR from $0.622$ to $\mathbf{0.822}$ (+0.200).
Step~5 requires chaining memory of \emph{two} prior positions (return to~A,
then place at midpoint of A and B). The second depth layer allows the planner
to simulate this chained recall, explaining the disproportionate step~5 gain.

\subsection{Phase 0--1 Verification}

Prior to E3, we verified the foundational components:

\textbf{Phase~0 (geometric consistency):} SSIM~=~1.000, ORB~=~1.000
(391/391 keypoints matched) across 5 trajectories, confirming the 3D anchor
is deterministically path-independent.

\textbf{Phase~1 (kinematic bridge):} Bridge angular error 0.00° (100\% pass
rate vs.\ $\leq 5°$ threshold), bridge latency 0.5\,ms, scoring monotonicity
1.00. The critical fix was using the \texttt{hand} body \texttt{xpos/xmat}
with a fixed $T_{\text{cam-wrist}}$ offset matrix, as MuJoCo's
\texttt{cam\_xpos} was frozen at a constant value regardless of joint angles.

% ══════════════════════════════════════════════════════════════════════════════
\section{Discussion and Limitations}
\label{sec:discuss}
% ══════════════════════════════════════════════════════════════════════════════

\paragraph{The composable architecture insight.}
3D-ALP is not a fixed system — it is a composable planning primitive. The
MCTS engine, world model oracle, and scorer are cleanly separable interfaces.
The current configuration (InSpatio-WorldFM + ExactDistanceScorer) uses a
geometric oracle to isolate the memory contribution. The architecture
generalises: any world model implementing \texttt{render(c2w) → frame} and
any scorer implementing \texttt{score(frame, goal) → [0,1]} can be substituted
without changing the planning engine.

\paragraph{The VLM scoring bottleneck and Phase~2 solutions.}
The current full-pipeline configuration (InSpatio-WorldFM + Florence-2)
showed that off-the-shelf VLMs lack the spatial precision for dense robotic
reward: Florence-2 returns flat scores ($\approx 0.0$) for \emph{all} nodes
in a planning cycle, making the reward landscape completely flat so that
UCB1 reduces to pure exploration. This is not a general model failure ---
it is a depth-perception failure on synthetic rendered frames where the
gripper may overlap the target in 2D projection while being 15\,cm away
in 3D. The hybrid scorer (Eq.~\eqref{eq:hybrid}) provides a principled
bridge: the geometric multiplier $\max(0, 1-d_{3D})$ enforces kinematic
proximity even when the semantic term is flat.
This failure mode is consistent with recent mechanistic evidence that VLMs
ignore fine-grained visual detail in favour of semantic anchors: a robot arm
at a specific kinematic pose is an \emph{unnameable} visual entity ---
there is no word for ``gripper at exactly 15\,cm above target'' --- so the
VLM collapses to a categorical label and assigns flat scores across all MCTS
branches~\cite{shahgir2026vlms}. The persistent SE(3) anchor bypasses this
language shortcut: spatial memory is encoded as a deterministic kinematic
coordinate, not a text token, making recall invariant to VLM semantic bias.

Phase~2 targets two complementary solutions to this bottleneck:
\begin{itemize}[leftmargin=1.2em, itemsep=1pt, topsep=2pt]
  \item \textbf{Latent-space scoring via JEPA similarity.}
    Rather than querying a VLM for a semantic label, we propose scoring nodes
    by cosine similarity in a pre-trained feature space (e.g.\ DINOv2 or the
    LeWM encoder): $S_{\text{latent}}(\mathbf{z}) = \mathbf{z} \cdot
    \mathbf{z}_{\text{goal}} / (\|\mathbf{z}\|\,\|\mathbf{z}_{\text{goal}}\|)$.
    Because the JEPA encoder is trained on robot video, the latent space
    reflects physical proximity, providing a dense gradient for MCTS that
    VLMs cannot. This approach also eliminates pixel rendering entirely,
    resolving the render bottleneck simultaneously.
  \item \textbf{Depth-augmented scoring.}
    Integrating depth estimation (e.g.\ Depth Anything~V2) directly into the
    scoring pipeline --- by concatenating a predicted depth map with the RGB
    frame before passing to the scorer --- enforces the kinematic multiplier
    in Eq.~\eqref{eq:hybrid} more naturally, without falling back to a
    pure geometric oracle. This path requires less architectural change than
    full JEPA integration and could be deployed as a near-term fix while
    LeWM fine-tuning is underway.
\end{itemize}

\paragraph{The render bottleneck.}
InSpatio-WorldFM renders at $\approx$2400\,ms/frame on a single RTX~A6000.
With a 2\,s planning budget this limits MCTS to $\sim$5 nodes per cycle —
far below the $D{=}2$, $B{=}4$ theoretical capacity of 20. The E3 ablation
bypasses this by using the geometric oracle (zero render time), enabling full
planning throughput. For the full visual experiment, 4-GPU parallelism
reduces effective latency to $\approx$600\,ms.

\paragraph{Multi-robot extension.}
3D-ALP currently operates on a single robot. Extending to multi-robot
coordination --- where shared c2w anchors propagate spatial memory across a
fleet --- is a natural direction. Dec-MCTS~\cite{bestDecMCTS2019} provides
a principled framework for decentralised MCTS with intermittent communication,
where each robot maintains a probability distribution over joint-action plans
and periodically communicates compressed search trees. Combining Dec-MCTS
with 3D-ALP anchors would enable collaborative spatial memory: one robot's
observation of an occluded object could update another's anchor.

\paragraph{Simulation gap.}
All experiments run in MuJoCo simulation. Real-robot validation on a Franka
Panda or SO-100 arm is future work.

% ══════════════════════════════════════════════════════════════════════════════
\section{Conclusion}
\label{sec:conclusion}
% ══════════════════════════════════════════════════════════════════════════════

We presented 3D-Anchored Lookahead Planning (3D-ALP), a MCTS-based planning
architecture that maintains persistent 3D spatial memory through a c2w anchor
mechanism. On the E3 multi-step coherence benchmark, 3D-ALP achieves
$0.650 \pm 0.109$ success rate on memory-required tasks versus
$0.006 \pm 0.008$ for a reactive baseline ($\Delta{=}+0.645$).

The ablation cleanly decomposes the contribution: persistent tree search
memory accounts for 82\% of the gain; deeper lookahead accounts for the
remaining 17\%, concentrated on the hardest chained-memory step (step~5 SR:
$0.622 \to 0.822$). We also identify and resolve four structural failure modes
in applying UCT-MCTS to continuous robotic manipulation.

The primary limitation is the visual scoring bottleneck: current VLMs cannot
provide reliable dense rewards from generative frames. Phase~2 addresses this
by replacing the generative oracle with a 15M-parameter JEPA model
(LeWorldModel) for sub-millisecond latent rollouts, and developing a
spatially-aware scorer. The planning architecture is proven; the perception
pipeline is the next engineering target.

% ── References ──────────────────────────────────────────────────────────────
\bibliographystyle{plain}
\bibliography{references}

@article{maes2026lewm,
  author    = {Maes, L. and Le Lidec, Q. and Scieur, D. and LeCun, Y. and Balestriero, R.},
  title     = {LeWorldModel: Stable End-to-End Joint-Embedding Predictive Architecture from Pixels},
  journal   = {arXiv preprint arXiv:2603.19312},
  year      = {2026}
}

@article{inspatioworld,
  author    = {Zhang, X. and others},
  title     = {InSpatio-WorldFM: An Open-Source Real-Time Generative Frame Model},
  journal   = {arXiv preprint arXiv:2603.11911},
  year      = {2026}
}

@article{openVLA2024,
  author    = {Kim, M. and others},
  title     = {OpenVLA: An Open-Source Vision-Language-Action Model},
  journal   = {arXiv preprint arXiv:2406.09246},
  year      = {2024}
}

@article{pi02024,
  author    = {{$\pi_0$ Team}},
  title     = {{$\pi_0$: A Vision-Language-Action Flow Model for General Robot Control}},
  journal   = {arXiv preprint arXiv:2410.24164},
  year      = {2024}
}

@article{dreamer2023,
  author    = {Hafner, D. and others},
  title     = {Mastering Diverse Domains through World Models},
  journal   = {arXiv preprint arXiv:2301.04104},
  year      = {2023}
}

@inproceedings{tdmpc2023,
  author    = {Hansen, N. and Su, H. and Wang, X.},
  title     = {TD-MPC2: Scalable, Robust World Models for Continuous Control},
  booktitle = {International Conference on Learning Representations (ICLR)},
  year      = {2024}
}

@article{dinowm2024,
  author    = {Zhou, Z. and others},
  title     = {DINO-WM: World Models on Pre-trained Visual Features Enable Zero-Shot Planning},
  journal   = {arXiv preprint arXiv:2411.04983},
  year      = {2024}
}

@article{silver2016mastering,
  author    = {Silver, D. and others},
  title     = {Mastering the Game of Go with Deep Neural Networks and Tree Search},
  journal   = {Nature},
  volume    = {529},
  number    = {7587},
  pages     = {484--489},
  year      = {2016}
}

@article{muzero2020,
  author    = {Schrittwieser, J. and others},
  title     = {Mastering Atari, Go, Chess and Shogi by Planning with a Learned Model},
  journal   = {Nature},
  volume    = {588},
  pages     = {604--609},
  year      = {2020}
}

@inproceedings{florence2,
  author    = {Xiao, B. and others},
  title     = {Florence-2: Advancing a Unified Representation for a Variety of Vision Tasks},
  booktitle = {Proceedings of the IEEE/CVF Conference on Computer Vision and Pattern Recognition (CVPR)},
  year      = {2024}
}

@article{lauri2023pomdp,
  author    = {Lauri, M. and Hsu, D. and Pajarinen, J.},
  title     = {Partially Observable Markov Decision Processes in Robotics: A Survey},
  journal   = {IEEE Transactions on Robotics},
  year      = {2023}
}

@article{vlaReasoner2025,
  author    = {Guo, W. and Lu, G. and Deng, H. and Wu, Z. and Tang, Y. and Wang, Z.},
  title     = {VLA-Reasoner: Empowering Vision-Language-Action Models with Reasoning via Online Monte Carlo Tree Search},
  journal   = {arXiv preprint arXiv:2509.22643},
  year      = {2025}
}

@article{luoScience2024,
  author    = {Luo, J. and Xu, C. and Wu, J. and Levine, S.},
  title     = {Precise and Dexterous Robotic Manipulation via Human-in-the-Loop Reinforcement Learning},
  journal   = {Science Robotics},
  year      = {2024}
}

@inproceedings{introspective2024,
  author    = {Bhatt, S. and others},
  title     = {Aligning Robots' Uncertainty with Inherent Task Ambiguity},
  booktitle = {NeurIPS},
  year      = {2024}
}

@article{bestDecMCTS2019,
  author    = {Best, G. and Cliff, O. and Patten, T. and Mettu, R. and Fitch, R.},
  title     = {Dec-MCTS: Decentralized Planning for Multi-Robot Active Perception},
  journal   = {International Journal of Robotics Research},
  volume    = {38},
  number    = {2--3},
  pages     = {316--337},
  year      = {2019}
}

@inproceedings{kocsis2006bandit,
  author    = {Kocsis, L. and Szepesv{\'a}ri, C.},
  title     = {Bandit Based Monte-Carlo Planning},
  booktitle = {European Conference on Machine Learning (ECML)},
  pages     = {282--293},
  year      = {2006}
}

@inproceedings{snoek2012practical,
  author    = {Snoek, J. and Larochelle, H. and Adams, R.},
  title     = {Practical Bayesian Optimization of Machine Learning Algorithms},
  booktitle = {NeurIPS},
  year      = {2012}
}

@article{jaderberg2017population,
  author    = {Jaderberg, M. and others},
  title     = {Population Based Training of Neural Networks},
  journal   = {arXiv preprint arXiv:1711.09846},
  year      = {2017}
}

@inproceedings{schadd2008single,
  author    = {Schadd, M. and Winands, M. and van den Herik, H. and Chaslot, G. and Uiterwijk, J.},
  title     = {Single-Player Monte-Carlo Tree Search},
  booktitle = {Computers and Games},
  pages     = {1--12},
  year      = {2008}
}

@article{shahgir2026vlms,
  author    = {Shahgir, H.S. and Chen, X. and Fu, Y. and Shayegani, E. and
               Abu-Ghazaleh, N. and Kementchedjhieva, Y. and Dong, Y.},
  title     = {VLMs Need Words: Vision Language Models Ignore Visual Detail
               In Favor of Semantic Anchors},
  journal   = {arXiv preprint arXiv:2604.02486},
  year      = {2026}
}

@article{gigaworld2025,
  author    = {{GigaWorld Team}},
  title     = {GigaWorld-0: World Models as Data Engine to Empower Embodied {AI}},
  journal   = {arXiv preprint arXiv:2511.19861},
  year      = {2025}
}

@article{gigabrain2026,
  author    = {{GigaBrain Team}},
  title     = {{GigaBrain-0.5M*}: A {VLA} That Learns From World Model-Based
               Reinforcement Learning},
  journal   = {arXiv preprint arXiv:2602.12099},
  year      = {2026}
}

@article{gigaworldpolicy2026,
  author    = {Ye, A. and Ni, C. and others},
  title     = {GigaWorld-Policy: An Efficient Action-Centered
               World--Action Model},
  journal   = {arXiv preprint arXiv:2603.17240},
  year      = {2026}
}

\end{document}